\documentclass[runningheads]{llncs}

\usepackage{eccv}

\usepackage{eccvabbrv}

\usepackage{graphicx}
\usepackage{booktabs}

\usepackage[accsupp]{axessibility}  

\usepackage{hyperref}

\usepackage{orcidlink}

\makeatletter
\newcommand{\corrautext}{Corresponding authors.}
\newcommand{\corrauthor}{\unskip$^{*}$}
\pretocmd{\maketitle}{%
  \protected@xdef\@thanks{%
    \@thanks
    \protect\footnotetext[1]{\corrautext}%
  }%
}{}{}
\makeatother

\usepackage{multirow}
\usepackage[table,xcdraw]{xcolor}
\usepackage{pifont}
\usepackage{algorithm}
\usepackage{algpseudocode}
\usepackage{amsmath, amssymb}
\usepackage{enumitem}
\usepackage{wrapfig}

\usepackage[most]{tcolorbox}
\newtcolorbox{principlebox}[1][]{
  enhanced,
  colback=gray!5,
  colframe=gray!60,
  boxrule=0.5pt,
  arc=2pt,
  left=6pt, right=6pt,
  top=4pt, bottom=4pt,
  title=#1,
  coltitle=black,
  fonttitle=\bfseries,
  fontupper=\scriptsize
}

\usepackage{fvextra}
\DefineVerbatimEnvironment{jsoncode}{Verbatim}
  {breaklines=true, fontsize=\scriptsize}

\begin{document}

\title{VDC-Agent: When Video Detailed Captioners Evolve Themselves via Agentic Self-Reflection}

\titlerunning{VDC-Agent}

\author{Qiang Wang\inst{1} \and
Xinyuan Gao\inst{2} \and
Yuhang He\inst{1}\corrauthor \and
Jizhou Han\inst{1} \and
Jiangyang Li\inst{1} \and
Songlin Dong\inst{3,4}\corrauthor \and
Zhiheng Ma\inst{3,4,5}\corrauthor \and
Yihong Gong\inst{1,3}\corrauthor}

\authorrunning{Q.~Wang et al.}


\institute{State Key Laboratory of Human-Machine Hybrid Augmented Intelligence, Institute of Artificial Intelligence and Robotics, Xi'an Jiaotong University \and
Kuaishou Technology \and
Shenzhen University of Advanced Technology \and
Guangdong Provincial Key Laboratory of Computility Microelectronics \and
Shenzhen Institutes of Advanced Technology, Chinese Academy of Sciences
\email{\{qwang,gxy010317\}@stu.xjtu.edu.cn,ygong@mail.xjtu.edu.cn}}


\maketitle


\begin{abstract}
Existing Video Detailed Captioning (VDC) methods predominantly rely on costly human annotations or distillation from powerful proprietary models, creating a dependency on external supervision. In this paper, we propose \textbf{VDC-Agent}, an autonomous self-evolving framework that empowers a single Multimodal Large Language Model (MLLM) to generate and refine high-quality captions through principle-guided self-reflection. To overcome the inference latency inherent in iterative refinement, we further propose to \textit{internalize} this reflective capability into the model. Specifically, we construct VDC-Agent-19K, a preference dataset derived from the agent's self-scored trajectories, and introduce a Curriculum Direct Preference Optimization (DPO) strategy. This strategy leverages the quality gap between generated candidates to progressively align the model from easy to hard samples. Extensive experiments demonstrate that VDC-Agent achieves state-of-the-art performance on VDC and DREAM-1K benchmarks, generating captions with superior detail and faithfulness. Crucially, our internalization strategy retains the inference efficiency of the base model while significantly enhancing its generalization capabilities, as validated by both quantitative metrics and human evaluation.
\par\addvspace\baselineskip
\noindent\textbf{Project Page:}\enspace\url{https://vdcagent.github.io}
\end{abstract}

\section{Introduction}
With the explosive growth of online videos, understanding complex visual and temporal information has become a core capability for modern multimodal AI systems. Within this landscape, \textbf{Video Detailed Captioning (VDC)}~\cite{auroracap,tarsier,tarsier2,videochat} serves as a fundamental task that aims to generate fine-grained and comprehensive descriptions capturing objects, actions, interactions, and scene transitions in videos. By enabling a more precise and detailed understanding of video content, VDC facilitates progress in a wide range of applications such as visual question answering~\cite{lei2018tvqa,zhong2022video}, open-vocabulary semantic segmentation~\cite{yin2025semi,yin2026depmatch,yin2026pca}, object detection~\cite{song2025learning,yin2026incremental}, text-to-video generation~\cite{hong2022cogvideo,sun2024sora}, video retrieval~\cite{gabeur2020multi}, and temporal localization~\cite{liu2024mllm,zhang2025weakly}.

\begin{figure}[t]
  \centering
  \includegraphics[width=0.93\textwidth]{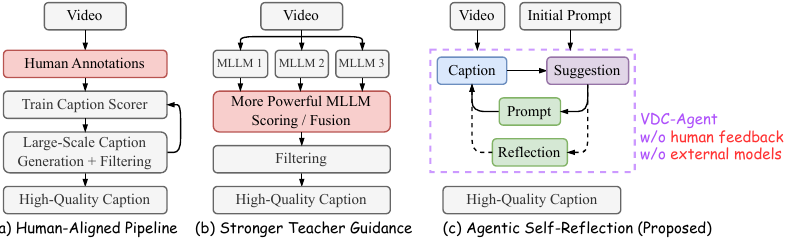}
  \caption{\textbf{Comparison of video captioning paradigms.}
  (a) Human-aligned pipelines rely on manual annotations to train caption scorers. 
  (b) Multi-MLLM-based pipelines depend on multiple or stronger MLLMs for scoring or fusion. 
  (c) Our proposed \textbf{VDC-Agent} achieves self-improvement through agentic self-reflection, requiring neither human annotations nor larger models.}
  \label{fig:intro_fig}
\end{figure}

Recent progress in VDC largely stems from fine-tuning Multimodal Large Language Models (MLLMs) on curated video-caption datasets~\cite{cockatiel,avc-dpo,videocap-r1}. While effective, these methods typically depend on external supervision, as shown in \cref{fig:intro_fig}~(a)(b). Approaches like ShareGPT4Video~\cite{sharegpt4video}, AVC-DPO~\cite{avc-dpo}, and VideoCap-R1~\cite{videocap-r1} rely on distilling capabilities from powerful proprietary or open-source models (\textit{e.g.}, GPT-4V~\cite{gpt4}, Qwen-72B~\cite{qwen2.5vl}), whereas Cockatiel~\cite{cockatiel} relies on extensive human annotations to align with preferences, and OwlCap~\cite{owlcap} leverages more powerful models to fuse captions from different MLLMs.

However, existing VDC paradigms are trapped in an unsustainable resource dilemma: they either suffer from prohibitive human annotation costs or rely heavily on the massive computation of larger teacher models. To break this deadlock, we ask: \textit{Can we enable a model to act as its own ``critic'' (akin to human introspection) to autonomously evaluate and correct its outputs without external intervention?} Achieving this would eliminate the dependency on expensive external MLLMs and extensive manual annotations. To this end, we propose transforming the captioner into an \textbf{autonomous agent}. By endowing it with intrinsic reflective capabilities, the agent establishes a closed loop to independently generate, evaluate, and refine its captions through iterative self-reflection.

To this end, we propose \textbf{VDC-Agent}, a self-evolving video captioning framework that enables MLLMs to improve themselves through iterative self-reflection without requiring stronger external supervision. As depicted in \cref{fig:intro_fig}~(c), VDC-Agent forms a closed-loop system that continuously refines its captioning ability by alternating between caption generation, evaluation, and prompt refinement. Given a collection of unlabeled videos, the model first generates captions using an initial prompt. It then conducts self-assessment based on a set of principles describing what constitutes a good caption (such as coverage of objects, actions, and temporal dynamics), assigns a quality score to the caption, and produces textual suggestions for improvement. These suggestions guide an internal \emph{prompt refiner}, which updates the prompt in the next iteration. If the newly generated caption is even worse than the previous one, the model triggers a \emph{self-reflection} mechanism that revisits the chain of thought used in the last prompt refinement, diagnosing why the previous update failed and avoiding the same mistake in subsequent steps. Through repeated cycles of caption–evaluation–refinement, our VDC-Agent can generate higher-quality video descriptions.

However, the iterative nature of VDC-Agent inevitably increases inference latency. To reconcile high performance with efficiency, we propose to \textit{internalize} the self-reflection capability into the MLLM, enabling it to achieve the effect of multiple rounds of reflection within a single inference step. Specifically, we apply VDC-Agent to a high-resolution video corpus (Cockatiel-4K~\cite{cockatiel}) to generate scored caption trajectories. By selecting the best and worst candidates, we construct a preference dataset, \textbf{VDC-Agent-19K}, comprising 18,886 pairs with quantified quality gaps. To fully exploit these pairs, we introduce a \textit{Curriculum Direct Preference Optimization (DPO)} strategy that progressively learns from easy (large score gap) to hard (small score gap) samples. We fine-tune Qwen2.5-VL-7B-Instruct with this strategy to obtain VDC-Agent-7B.

Extensive experiments validate the effectiveness of our framework from multiple perspectives. On the VDC benchmark, our model achieves state-of-the-art results (49.08\% accuracy, 2.50 score), demonstrating its capacity to generate richer and more fine-grained details. On the DREAM-1K benchmark, the significant improvements in Precision and Recall (37.4\% F1) suggest that our method produces \textit{faithful descriptions rather than merely longer text}. Furthermore, human evaluation verifies that the positive-negative pairs in VDC-Agent-19K are \textit{highly consistent with human preferences}. Finally, ablation studies demonstrate that while the self-reflection mechanism is effective, internalizing it yields the best of both worlds: it \textit{further improves accuracy and generalization} compared to test-time iteration while drastically \textit{reducing inference time} to the level of the base model.

Our contributions can be summarized as follows:
\begin{itemize}
    \item We propose VDC-Agent, an autonomous framework that enables a single MLLM to generate and refine detailed video captions via self-reflection, eliminating the need for human annotation or external teacher models.
    \item We construct a preference dataset VDC-Agent-19K and introduce curriculum DPO that exploits the score gap between preferred and dispreferred captions as a difficulty signal to enable easy-to-hard preference alignment.
    \item We achieve state-of-the-art performance on two VDC benchmarks. Human evaluation and ablation studies validate the effectiveness and efficiency of our self-reflection and internalization strategies.
\end{itemize}

\section{Related Work}
\vspace{-1mm}
\subsection{Video Detailed Captioning}
Video detailed captioning (VDC)~\cite{auroracap,videochat,qasim2025dense,lian2025describe,wei2025longcaptioning,chen2025vidcapbench,pllava,islam2024video} is a fundamental task in video understanding that aims to generate precise and comprehensive descriptions of video content. Early non-LLM approaches~\cite{krishnamoorthy2013generating,guadarrama2013youtube2text} typically produced short, fragmentary captions with limited temporal grounding, which made them hard to use in downstream reasoning or retrieval systems. With the rapid progress of Multimodal Large Language Models (MLLMs)~\cite{qwen2.5vl,internvl2.5,dong2025beyond,ding2024lobg}, recent works have substantially improved the fluency and coherence of video captions. However, generic MLLMs are not explicitly optimized for fine-grained and temporally grounded descriptions, and they often miss subtle events, camera operations, or background cues that are critical for VDC.

Recent methods construct high-quality video-caption pairs and then fine-tune via LoRA~\cite{zhao2026learning} or Adapter~\cite{wang2024non,li2025dynamic} on these curated datasets. Cockatiel~\cite{cockatiel} and Vriptor~\cite{vriptor} build dense, human-annotated resources or human-preferenced scorers to obtain detailed captions, while ShareGPT4Video~\cite{sharegpt4video}, Shot2Story~\cite{shot2story}, and LLaVA-Video~\cite{llava-video} leverage GPT-4V or GPT-4o to synthesize large-scale annotations that are further refined by filtering or human checks. More recently, AVC-DPO~\cite{avc-dpo} and VideoCap-R1~\cite{videocap-r1} employ powerful open-source models (\emph{e.g.}, Qwen-72B) as teachers or reward providers to train smaller captioners with preference optimization or reinforcement learning. Despite their success, these approaches still hinge on stronger MLLMs or extensive human supervision for data construction, which leads to considerable annotation and computation costs and limits scalability and reproducibility in practice.

\subsection{Multimodal Large Language Model Agent}
Recent research has shifted from viewing LLMs as passive predictors to agents that interleave reasoning with actions~\cite{react,reflexion,self-refine,llm_agent_survey,yuan2024self,yu2025rlaif,han2026consistent,han2026learning,bai2022constitutional,self-instruct,spin,wang2023promptagent}. These frameworks enable robust decision-making through iterative ``think-then-act'' cycles. In the multimodal domain, agents extend this capability by integrating vision experts~\cite{mmreact}, navigating GUIs~\cite{cogagent,appagent,li2026trajectory}, or querying long videos~\cite{videoagent1,videoagent2}. Distinct from these approaches that interact with external environments or tools, our VDC-Agent internalizes the action space. It operates directly on its own prompts and captions, establishing a closed loop of generation, principle-guided evaluation, and self-reflection to autonomously enhance video detailed captioning without relying on external teacher models.

\begin{figure*}[t]
  \centering
  \includegraphics[width=0.95\textwidth]{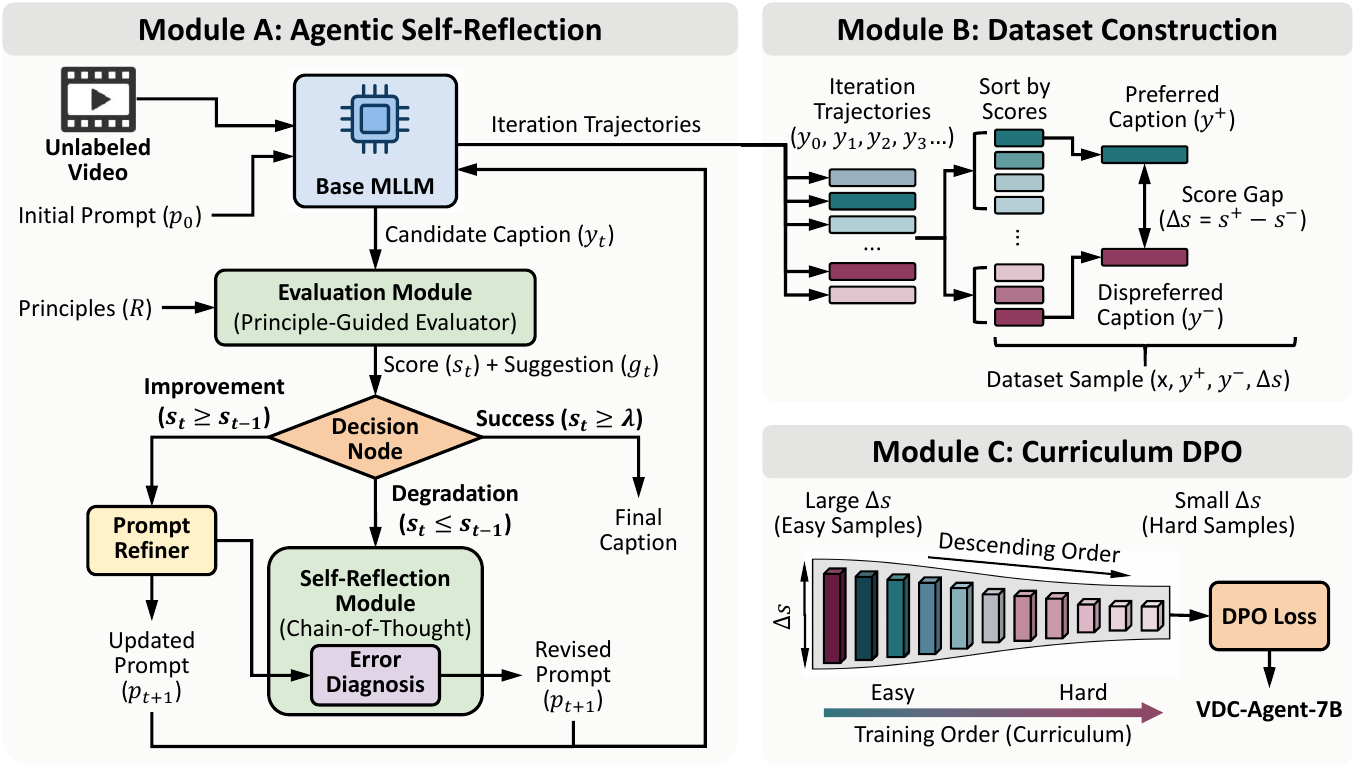}
  \caption{\textbf{Overview of the VDC-Agent Framework.} 
  \textbf{(A) Agentic Self-Reflection:} The agent iteratively refines captions through a closed-loop system of generation, evaluation, and self-reflection, dynamically handling both improvements and degradations.
  \textbf{(B) Dataset Construction:} We extract preference pairs ($y^+, y^-$) from the iteration trajectories based on quality scores to form the dataset.
  \textbf{(C) Curriculum DPO:} The model is fine-tuned by organizing these pairs from easy (large score gap $\Delta s$) to hard (small $\Delta s$), effectively internalizing the reflective capability.}
  \label{fig:framework}
\end{figure*}

\vspace{-1mm}
\section{Method}\label{sec:method}
To obtain a stronger video detailed captioner, we first construct high–quality / low–quality caption pairs with an agentic data generator, \textbf{VDC-Agent}, and then fine-tune a multimodal large language model (MLLM) using direct preference optimization (DPO). \cref{sec:3.1} details how VDC-Agent iteratively improves caption quality. \cref{sec:3.2} shows how the resulting sequence of captions is transformed into a training dataset. Finally, \cref{sec:3.3} explains how the positive–negative pairs are leveraged to train our model via curriculum DPO.

\subsection{VDC-Agent}\label{sec:3.1}
\cref{fig:framework} illustrates the overall framework of our VDC-Agent. Given a collection of videos, VDC-Agent automatically generates multiple captions for each video, scores their quality, and finally produces preference pairs used for DPO training. The key idea is to let an MLLM iteratively refine the task prompts via self-reflection, so that later iterations yield more detailed captions.

We set a maximum number of iterations $T$ and use $t \in \{0,1,\dots,T\}$ to index the interaction steps. At step $t$, the prompt, caption, score, suggestion are denoted by $p_t$, $y_t$, $s_t$, and $g_t$, respectively. As preparation, the user only needs to specify an initial prompt $p_0$ (\emph{e.g.}, ``Please provide a detailed description of this video.'') and a set of textual principles $R$ that describe what a good caption should look like (\emph{e.g.}, covering camera motion, background, main objects, \emph{etc.}). The details of principles will be provided in the Appendix. After that, VDC-Agent can run fully automatically over a video collection.

\noindent\textbf{Caption generation.}
For a video $x$ and a given prompt $p_t$, a base MLLM produces a candidate caption:
\begin{equation}
  y_t = f(x;\Theta, p_t), \quad 0 \le t \le T ,
\end{equation}
where $f(\cdot;\cdot)$ denotes the base MLLM and $\Theta$ is its parameter set. When $t=0$, the caption $y_0$ is generated using the initial prompt $p_0$.

\noindent\textbf{Scoring and suggestion generation.}
The caption $y_t$ is then fed into the MLLM (which shares the same backbone but is used with different instructions) together with the video $x$ and the principle $R$. This \emph{principle-guided} MLLM outputs both a scalar quality score and natural-language suggestions for improving the prompt:
\begin{equation}
  (s_t, g_t) = f(x, y_t;\,\Theta, R), \quad 0 \le t \le T ,
\end{equation}
where $s_t$ is the quality score from 0 to 100, and $g_t$ is a textual suggestion describing how the next prompt should be revised (\emph{e.g.}, asking for more spatial details or emphasizing the main object).

\noindent\textbf{Iterative prompt refinement with reflection.}
VDC-Agent decides whether and how to update the prompt $p_t$ based on the score $s_t$. Given a score threshold $\lambda$, if the caption already satisfies the requirement, \emph{i.e.}, $s_t \ge \lambda$, the agent stops and outputs the current caption. Otherwise, it refines the prompt and continues.

We distinguish two cases for unsuccessful captions. If the score increases or remains unchanged compared with the previous step, the agent simply performs \emph{prompt refinement}, using the suggestion $g_t$ as guidance. If the score drops below the previous score, this indicates that the last refinement was harmful. In this case VDC-Agent triggers a \emph{self-reflection} stage: it asks the MLLM to reason about why the previous refinement failed, using the current prompt and caption, with the previous round’s prompt–caption pair and the chain-of-thought (CoT) produced by the previous prompt refiner, and then proposes a more reliable prompt update. Formally, the prompt update rule can be written as:
\begin{equation}
p_{t+1} =
\begin{cases}
p_t, & s_t \ge \lambda,\\[2pt]
f(x, y_t, s_t;\,\Theta, p_{\mathrm{refine}}), & s_t \in [s_{t-1}, \lambda),\\[2pt]
f(x, y_t, s_t, p_t;\,\Theta, p_{\mathrm{reflect}}), & s_t < s_{t-1},
\end{cases}
\label{eq:prompt_update}
\end{equation}
where $p_{\mathrm{refine}}$ is the system-level instruction used by the prompt refiner and $p_{\mathrm{reflect}}$ is the instruction for the self-reflection module. 
The system prompts of the (self-reflection) prompt refiner will be provided in the Appendix. 
The self-reflection trigger rate and examples for generation will be provided in the Appendix.
For clarity, we omit the explicit CoT variables in \cref{eq:prompt_update}; in practice, the CoT generated in the previous step is provided as additional context so that the agent can diagnose errors and avoid repeating them. The detailed pseudo-code and algorithm are presented in the Appendix.
During the above process, each iteration produces a pair $(y_t, s_t)$. After the loop terminates due to either reaching the threshold or hitting the maximum iteration $T$, we keep all caption-score pairs generated for this video and pass them to the data construction stage.

\subsection{Dataset Construction by VDC-Agent}\label{sec:3.2}
High-quality video data is crucial for training reliable detailed captioners. We adopt the Cockatiel-4K~\cite{cockatiel} corpus, which contains 4,008 high-resolution videos sampled from OpenVid-1M. To expose the model to complementary aspects of video understanding, we follow the taxonomy in VDC~\cite{auroracap} and generate captions along five task dimensions: camera (shot type and camera motion), short (a concise summary), background (scene layout and context), main object (key objects and their attributes), and detailed (fine-grained temporal and spatial events). For each video $x$ and each task dimension $d$, VDC-Agent executes the iterative procedure of \cref{sec:3.1} and produces a trajectory of caption-score pairs:
\begin{equation}
\mathcal{P}(x,d) = \{(y_t, s_t)\}_{t=0}^{T_v(x,d)}, \quad 1 \le T_v(x,d) \le T,
\end{equation}
where $y_t$ is the caption at iteration $t$, $s_t$ is its quality score, and $T_v(x,d)$ is the number of iterations before termination by threshold or by the cap $T$. Across 4,008 videos and 5 dimensions, we obtain a total of 20,040 raw caption sets.

\noindent\textbf{Rule-based filtering.}
If a video already attains a score no smaller than the threshold $\lambda$ in the first iteration (\emph{i.e.}, $\lvert \mathcal{P}(x,d) \rvert = 1$), the existing MLLM provides a sufficiently good caption without any refinement. Such cases offer little learning signal for preference modeling and are therefore removed. This filter discards 1,078 $(x,d)$ pairs. In addition, we remove 76 pairs due to JSON formatting / parsing errors during automatic generation. All filtering is \emph{fully automatic} and rule-based, \emph{requiring no human annotation}. After filtering, we retain 18,886 caption sets for the next stage. The details of the dataset and the filtered data will be provided in the Appendix.

\noindent\textbf{Constructing positive-negative pairs.}
For every remaining set $\mathcal{P}(x,d)$, we sort captions by their scores and select the highest-scoring caption as the \emph{best} caption $y^{+}$ with score $s^{+}$, and the lowest-scoring caption as the \emph{worst} caption $y^{-}$ with score $s^{-}$. To quantify the preference strength (and later drive the curriculum), we compute the score gap $\Delta s = s^{+} - s^{-}$. Each video and task dimension then contributes one training tuple $(x,\; y^{+},\; y^{-},\; \Delta s)$, forming the dataset used by the training in \cref{sec:3.3}.

\subsection{VDC Learning with Curriculum DPO}\label{sec:3.3}
We fine-tune the base MLLM with DPO, which updates a policy $\pi_\theta$ from paired preferences without training a reward model or running on-policy RL.

\noindent\textbf{Vanilla DPO.}
Given a video $x$, a preferred caption $y^{+}$ and a dispreferred caption $y^{-}$, DPO~\cite{dpo} maximizes the probability that $\pi_\theta$ prefers $y^{+}$ over $y^{-}$ while keeping it close to a reference policy $\pi_{\mathrm{ref}}$:
\begin{equation}
\label{eq:dpo}
\mathcal{L}_{\text{DPO}}(\theta; x, y^{+}, y^{-}) = 
- \log \sigma\!\left(
\beta \Big[
\log \frac{\pi_{\theta}(y^{+}\!\mid x)}{\pi_{\theta}(y^{-}\!\mid x)}
- 
\log \frac{\pi_{\mathrm{ref}}(y^{+}\!\mid x)}{\pi_{\mathrm{ref}}(y^{-}\!\mid x)}
\Big] \right),
\end{equation}
where $\beta>0$ controls the implicit KL strength.

\noindent\textbf{Why curriculum?}
Vanilla DPO treats every pair equally, which is suboptimal for our data. In our setting, pair difficulty varies: some preferred-dispreferred pairs differ markedly (large semantic gap), which are ideal for early-stage coarse adaptation~\cite{bengio2009curriculum,wang2025dualcp,han2026goal}; others are much closer (small gap), which are valuable for late-stage fine-grained alignment~\cite{han2025learn,li2026parameter,ding2025class}. Fortunately, the dataset produced by VDC-Agent naturally provides a difficulty signal, \emph{i.e.}, the score gap $\Delta s$, quantifying how much better the preferred caption is than the dispreferred one.

\noindent\textbf{Curriculum DPO.}
To let the model learn \emph{from easy to hard} in an optimization-friendly manner, we adopt a simple yet effective curriculum DPO~\cite{bengio2009curriculum,currdpo}. We exploit the score gap $\Delta s$ as a built-in difficulty signal and only modify the sampling order. Specifically, we sort all tuples $\mathcal{D}=\{(x_i,y_i^{+},y_i^{-},\Delta s_i)\}_{i=1}^{N}$ by $\Delta s_i$ in descending order ($\Delta s_1 \ge \cdots \ge \Delta s_N$), and feed mini-batches sequentially along this list (large $\Delta s \!\rightarrow$ small $\Delta s$). The optimization objective is
\begin{equation}
  \min_{\theta} \; \sum_{i=1}^{N} \mathcal{L}_{\text{DPO}}(\theta; x_i, y_i^{+}, y_i^{-}),
\ \text{with order } i=1\!\rightarrow\!N.
\end{equation}
Intuitively, large-gap pairs yield strong, low-variance gradients that quickly steer the policy toward the correct global behavior; small-gap pairs then refine subtle distinctions without delicate weight tuning. Moreover, a cosine-decay learning-rate schedule naturally complements this curriculum: early confident pairs are learned with larger step sizes, while later ambiguous pairs are absorbed with smaller steps. Empirically, we find that this simple curriculum accelerates convergence and yields better final alignment for video detailed captioning.

\section{Experiments}
\subsection{Experimental Setup}
\noindent\textbf{Training data.}
We construct our training dataset based on Cockatiel-4K~\cite{cockatiel} due to its high visual quality, which contains 4,008 high-resolution videos sampled from OpenVid-1M~\cite{openvid1m}. We only use the videos from Cockatiel-4K without captions. Importantly, there is no overlap between the training videos and those used in the VDC evaluation benchmarks. Following our VDC-Agent pipeline, we generate raw caption trajectories on five task dimensions (camera, short, background, main object, detailed), producing 20,040 raw sets in total. Applying the rule-based filtering in \cref{sec:3.2} to remove trivial one-step cases and malformed JSON outputs, we obtain 18,886 preference pairs for training, denoted as VDC-Agent-19K.

\noindent\textbf{Implementation details.}
We use Qwen2.5-VL-7B-Instruct~\cite{qwen2.5vl} as our base MLLM. We fine-tune with curriculum DPO using VDC-Agent-19K. For efficiency, only the LLM backbone is adapted with LoRA~\cite{lora,gao2024beyond,wang2025boosting} while all other parameters are frozen. We set the LoRA rank and alpha to 16 and 32, respectively, with dropout 0.1. Training runs for 3 epochs with a cosine-decay learning-rate schedule (initial LR $5\times10^{-5}$, warmup $10\%$ of total steps). We train on $4\times$ NVIDIA A800 GPUs with a global batch size of 16. The maximum number of iterations $T$ is set to 4. $\lambda$ is set to 90. Following AVC-DPO~\cite{avc-dpo}, we set the maximum number of sampled frames to 32 and tokens per frame to 128. For decoding, we use a beam size of 1, a temperature of 0.1, and a maximum caption length of 1024.

\noindent\textbf{Evaluation benchmarks.} We evaluate our model on the VDC and DREAM-1K benchmarks to assess the models in video captioning tasks. VDC is a detailed video captioning benchmark containing 1,027 videos that span a wide range of categories, and we report the VDCscore following AuroraCap~\cite{auroracap}. DREAM-1K is a collection of 1,000 annotated video clips with diverse complexities, for which we report the event-level precision, recall, and F1 score following Tarsier~\cite{tarsier}.

\begin{table*}[t]
\fontsize{6pt}{7pt}\selectfont
\centering
\setlength{\tabcolsep}{0.8pt}
\caption{\textbf{VDCscore comparison on the VDC benchmark.} ``\textbf{H}uman'' indicates whether human annotations are used. ``\textbf{E}xternal'' indicates whether larger external teacher models are involved. We report Accuracy/Score for five dimensions (higher is better) and the average across dimensions. Qwen2.5-VL-7B-Instruct is the baseline model of our VDC-Agent-7B. \textbf{Bold} denotes the best results, and \underline{underline} denotes the second-best results.}
\begin{tabular}{l|cc|ccccc|c}
\toprule
\multirow{2}{*}{Model} & \multirow{2}{*}{H} & \multirow{2}{*}{E} & Camera & Short  & Background   & Main Object  & Detailed     & Average      \\
    &  &  & Acc./Score    & Acc./Score    & Acc./Score    & Acc./Score    & Acc./Score    & Acc./Score    \\ \midrule
\textbf{\textit{General MLLMs}} &  &  &  &  &  &  &  &  \\
Llama 3.1-8B~\cite{llama3}                   & - & - & 17.83/1.00   & 17.90/1.02   & 19.52/1.00   & 19.57/1.10   & 20.10/1.22   & 18.98/1.07   \\
Gemini 1.5 Pro~\cite{gemini1.5}              & - & - & 38.68/2.05   & 35.71/1.85   & 43.84/2.23   & 47.32/2.41   & 43.11/2.22   & 41.73/2.15   \\
GPT-5.4~\cite{gpt5.4}                                     & - & - & 46.35/2.38   & 35.56/1.82   & 47.04/2.40   & 44.65/2.28   & 45.09/2.30   & 43.74/2.24   \\
Gemini 3.1 Pro~\cite{gemini3.1pro}                              & - & - & 49.25/2.51   & 37.63/1.91   & 48.10/2.45   & 48.71/2.49   & 47.31/2.40   & 46.20/2.35   \\
LLaMA-VID-7B~\cite{llama-vid}                & - & - & 39.47/2.10   & 29.92/1.56   & 28.01/1.45   & 31.24/1.59   & 25.67/1.38   & 30.86/1.62   \\
Video-ChatGPT-7B~\cite{video-chatgpt}        & - & - & 37.46/2.00   & 29.36/1.56   & 33.68/1.70   & 30.47/1.60   & 24.61/1.26   & 31.12/1.62   \\
Video-LLaVA-7B~\cite{video-llava}            & - & - & 37.48/1.97   & 30.67/1.63   & 32.50/1.70   & 36.01/1.85   & 27.36/1.43   & 32.80/1.72   \\
VideoChat-Flash-7B~\cite{videochat-flash}    & - & - & 43.70/2.30   & 33.70/1.70   & 45.10/2.30   & 47.60/2.40   & 44.50/2.30   & 42.92/2.20   \\
Video-R1-7B~\cite{video-r1}                  & - & - & 42.70/2.20   & \underline{44.50/2.30} & 40.60/2.10   & 45.90/2.30   & 45.60/2.40   & 43.86/2.26   \\
Qwen3-VL-8B~\cite{qwen3vl}                                 & - & - & 48.44/2.45   & 36.58/1.86   & 47.69/2.42   & 46.94/2.41   & 46.75/2.39   & 45.28/2.31   \\
\midrule
\textbf{\textit{Video Caption MLLMs}} &  &  &  &  &  &  &  &  \\
ShareGPT4Video-8B~\cite{sharegpt4video}      & \ding{55} & \ding{51} & 33.28/1.76   & 39.08/1.94   & 35.77/1.81   & 37.12/1.89   & 35.62/1.84   & 36.17/1.85   \\
Vriptor~\cite{vriptor}                       & \ding{51} & \ding{55} & 37.64/1.96   & 38.35/2.00   & 37.11/1.94   & 37.02/1.93   & 38.49/2.00   & 37.72/1.97   \\
AuroraCap-7B~\cite{auroracap}                & \ding{55} & \ding{55} & 43.50/2.27   & 32.07/1.68   & 35.92/1.84   & 39.02/1.97   & 41.30/2.15   & 38.36/1.98   \\
Cockatiel-8B~\cite{cockatiel}                & \ding{51} & \ding{55} & 42.25/2.19   & 44.01/2.27   & 43.89/2.26   & 43.85/2.26   & 44.00/2.27   & 43.60/2.25   \\
VideoCap-R1-7B~\cite{videocap-r1}            & \ding{55} & \ding{51} & 41.70/2.30   & 35.20/1.90   & 47.20/2.50   & 47.00/2.50   & 43.80/2.40   & 42.98/2.32   \\
SynPO~\cite{synpo}                           & \ding{55} & \ding{55} & -/1.78 & -/1.94 & -/1.91 & -/1.87 & -/2.04 & -/1.91 \\
AVC-DPO-7B~\cite{avc-dpo}                    & \ding{55} & \ding{51} & \underline{50.40/2.66}   & 39.00/2.03   & \underline{49.90/2.57}   & \underline{50.50/2.58}   & \underline{48.90/2.54}   & \underline{47.70/2.47}   \\
OwlCap-7B~\cite{owlcap}                      & \ding{55} & \ding{51} & 46.90/2.50 & \textbf{47.00/2.40} & 46.50/2.40 & 47.10/2.50  & 46.80/2.40 & 46.90/2.40  \\ \midrule
LLaVA-OneVision-7B~\cite{llava-onevision}           & - & - &                 37.82/2.02 & 32.58/1.70 & 37.43/1.92 & 38.21/1.96  & 41.20/2.13 & 37.45/1.95 \\
\rowcolor[HTML]{EFEFEF} 
\quad w/ VDC-Agent (Ours)    & \ding{55} & \ding{55} & 44.15/2.31 & 35.22/1.84 & 45.02/2.26 & 43.55/2.25  & 47.96/2.37 & 43.18/2.21 \\
\rowcolor[HTML]{DFDFDF} 
\quad Improvement            & \ding{55} & \ding{55} & {\tiny +6.33/+0.29} & {\tiny +2.64/+0.14} & {\tiny +7.59/+0.34} & {\tiny +5.34/+0.29} & {\tiny +6.76/+0.24} & {\tiny +5.73/+0.26} \\

InternVL2.5-8B~\cite{internvl2.5}               & - & - &                 40.22/2.16 & 35.59/1.86 & 38.93/2.03 & 45.70/2.36  & 36.75/2.01 & 39.44/2.08 \\
\rowcolor[HTML]{EFEFEF} 
\quad w/ VDC-Agent (Ours)    & \ding{55} & \ding{55} & 48.08/2.53 & 38.17/1.96 & 46.59/2.32 & 49.13/2.48  & 42.43/2.23 & 44.88/2.30 \\
\rowcolor[HTML]{DFDFDF} 
\quad Improvement            & \ding{55} & \ding{55} & {\tiny +7.86/+0.37} & {\tiny +2.58/+0.10} & {\tiny +7.66/+0.29} & {\tiny +3.43/+0.12} & {\tiny +5.68/+0.22} & {\tiny +5.44/+0.22} \\

Qwen2.5-VL-7B~\cite{qwen2.5vl}                & - & - &                 42.61/2.18 & 37.76/1.91 & 44.10/2.22 & 49.00/2.47  & 46.25/2.35 & 43.95/2.23 \\
\rowcolor[HTML]{EFEFEF} 
\quad w/ VDC-Agent (Ours)    & \ding{55} & \ding{55} & \textbf{50.52/2.67} & 39.49/1.99 & \textbf{51.93/2.62} & \textbf{53.23/2.65}  & \textbf{50.21/2.55} & \textbf{49.08/2.50} \\ 
\rowcolor[HTML]{DFDFDF} 
\quad Improvement            & \ding{55} & \ding{55} & {\tiny +7.91/+0.49} & {\tiny +1.73/+0.08} & {\tiny +7.83/+0.40} & {\tiny +4.23/+0.18} & {\tiny +3.96/+0.20} & {\tiny +5.13/+0.27} \\
\bottomrule
\end{tabular}
\label{tab:vdcscore}
\end{table*}

\begin{table}[t]
\scriptsize
\centering
\setlength{\tabcolsep}{3pt}
\caption{\textbf{Comparison on the DREAM-1K benchmark.} ``Human'' indicates whether human annotations are used. ``External'' indicates whether larger external teacher models are involved. \textbf{Bold}: the best results. \underline{Underline}: the second-best results.}
\begin{tabular}{lccccc}
\toprule
Model                                     & Human         & External         & F1   & Precision & Recall \\
\midrule
\textbf{\textit{General MLLMs}}                             &           &           &      &           &        \\
Gemini 1.5 Pro~\cite{gemini1.5}           & -         & -         & 36.2 & 37.6      & 34.8   \\
GPT-5.4~\cite{gpt5.4}                      & -         & -         & 34.9 & 36.2      & 33.7   \\
Gemini 3.1 Pro~\cite{gemini3.1pro}         & -         & -         & 36.4 & 38.1      & \underline{34.9}   \\
LLaVA-Video-7B~\cite{llava-video}         & -         & -         & 32.5 & 37.9      & 28.4   \\
Video-R1-7B~\cite{video-r1}               & -         & -         & 33.3 & 32.2      & 34.5   \\
Qwen3-VL-8B~\cite{qwen3vl}                 & -         & -         & 32.5 & 33.6      & 31.5   \\
\midrule
\textbf{\textit{Video Caption MLLMs}}                       &           &           &      &           &        \\
Tarsier-7B~\cite{tarsier}                 & \ding{51} & \ding{51} & 34.6 & \underline{40.3}      & 30.2   \\
ShareGPT4Video-8B~\cite{sharegpt4video}   & \ding{55} & \ding{51} & 20.4 & 27.6      & 16.1   \\
Vriptor~\cite{vriptor}                    & \ding{51} & \ding{55} & 24.4 & 23.6      & 25.1   \\
AuroraCap-7B~\cite{auroracap}             & \ding{55} & \ding{55} & 20.8 & 24.4      & 18.1   \\
VideoCap-R1-7B~\cite{videocap-r1}         & \ding{55} & \ding{51} & 34.2 & 33.6      & 34.7   \\
OwlCap-7B~\cite{owlcap}                   & \ding{55} & \ding{51} & 34.7 & 34.1      & 35.3   \\
\midrule
LLaVA-OneVision-7B~\cite{llava-onevision} & -         & -         & 31.7 & 34.3      & 29.4   \\
\rowcolor[HTML]{EFEFEF} 
\quad w/ VDC-Agent (Ours)                 & \ding{55} & \ding{55} & \underline{36.9} (+5.2) & 39.6 (+5.3)      & 34.5 (+5.1)   \\
InternVL2.5-8B~\cite{internvl2.5}         & -         & -         & 27.6 & 34.7      & 22.9   \\
\rowcolor[HTML]{EFEFEF} 
\quad w/ VDC-Agent (Ours)                 & \ding{55} & \ding{55} & 33.9 (+6.3) & \textbf{40.5} (+5.8)      & 29.1 (+6.2)   \\
Qwen2.5-VL-7B~\cite{qwen2.5vl}           & -         & -         & 30.1 & 30.5      & 29.7   \\
\rowcolor[HTML]{EFEFEF} 
\quad w/ VDC-Agent (Ours)                 & \ding{55} & \ding{55} & \textbf{37.4} (+7.3) & 38.5 (+8.0)      & \textbf{36.3} (+6.6)   \\
\bottomrule
\end{tabular}
\label{tab:dream1k}
\end{table}

\begin{table}[t]
    \centering
    \begin{minipage}[c]{0.63\textwidth}
        \scriptsize
        \centering
        \captionof{table}{\textbf{Preference evaluation} on 500 randomly sampled caption pairs from VDC-Agent-19K. $y^{+}$ denotes the caption with a higher score, and $y^{-}$ denotes the caption with a lower score. For GPT-5.1, we report results for both input orders to analyze position bias, along with their average.}
        \label{tab:human_gpt_eval}
        \begin{tabular}{lcccc}
            \toprule
            \multirow{2}{*}{Evaluator} & \multirow{2}{*}{Input Order} & \multicolumn{3}{c}{Preference Distribution (\%)} \\
            \cmidrule(lr){3-5}
             &  & $y^{+}$ Better & Tie & $y^{-}$ Better \\
            \midrule
            Human Evaluator 1 & - & 75.8 & 17.0 & 7.2 \\
            Human Evaluator 2 & - & 69.0 & 24.6 & 6.4 \\
            Human Evaluator 3 & - & 67.6 & 22.2 & 10.2 \\
            \midrule
            \multirow{3}{*}{GPT-5.1} & $y^{-}$ then $y^{+}$ & 84.4 & 4.2 & 11.4 \\
                                     & $y^{+}$ then $y^{-}$ & 95.2 & 3.6 & 1.2 \\
                                     & Average              & 89.8 & 3.9 & 6.3 \\
            \bottomrule
        \end{tabular}
    \end{minipage}
    \hfill
    \begin{minipage}[c]{0.35\textwidth}
        \centering
        \includegraphics[width=\linewidth]{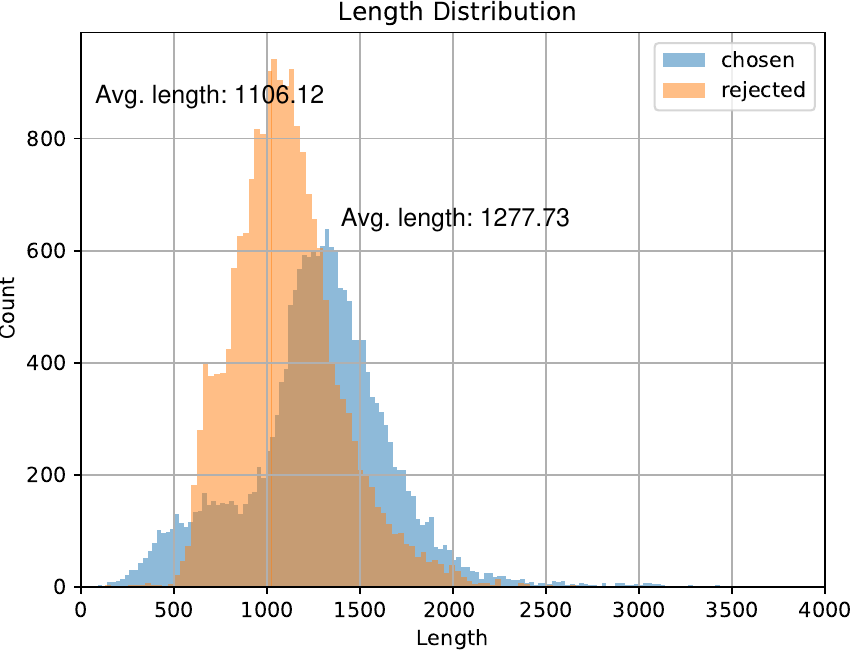}
        \captionof{figure}{Length distribution comparison between $y^{+}$ and $y^{-}$. The considerable overlap between the two distributions suggests that the preference for $y^{+}$ is not primarily driven by length.}
        \label{fig:length_distribution}
    \end{minipage}
\end{table}

\subsection{Main Results}
\noindent\textbf{Compared methods.}
We compare our method with two categories of models: General MLLMs, such as Llama 3.1-8B, and specialized Video Caption MLLMs with approximately 7B parameters, such as VideoCap-R1-7B, AVC-DPO-7B, and OwlCap-7B. Notably, while several top-performing baselines heavily rely on human annotations or distillation from larger teacher models (e.g., 72B parameters), our VDC-Agent operates entirely autonomously. To demonstrate the method-agnostic generalization of our framework, we apply it to three distinct base models: LLaVA-OneVision-7B, InternVL2.5-8B, and Qwen2.5-VL-7B.

\noindent\textbf{Evaluation on VDC benchmark.}
As shown in \cref{tab:vdcscore}, VDC-Agent establishes a new state-of-the-art on the VDC benchmark. Built on Qwen2.5-VL-7B, our model achieves the highest average accuracy (49.08\%) and score (2.50), outperforming strong teacher-dependent models like AVC-DPO-7B and OwlCap-7B. Crucially, our framework demonstrates robust cross-model transferability: applying VDC-Agent yields substantial and consistent improvements across all three base models (e.g., +5.73\% average accuracy for LLaVA-OneVision and +5.44\% for InternVL2.5). Dimension-wise, the gains are particularly pronounced in detail-oriented aspects such as camera (+7.91\%), background (+7.83\%), and main object (+4.23\%). In contrast, the improvement on the `Short' dimension is relatively modest (+1.73\%), because base MLLMs already possess strong global summarization capabilities, whereas our method primarily targets the refinement of visual details. This proves that agentic self-reflection effectively enhances fine-grained spatial and temporal video understanding without requiring external supervision.

\noindent\textbf{Evaluation on DREAM-1K benchmark.}
To verify that our performance gains reflect genuine improvements in visual grounding rather than overfitting to the VDCscore metric, we further evaluate on the DREAM-1K benchmark in \cref{tab:dream1k}. VDC-Agent consistently boosts performance across all evaluated backbones. Specifically, our Qwen2.5-VL-7B variant achieves the highest F1 score (37.4\%) and Recall (36.3\%) among all models, surpassing strong baselines like Tarsier-7B and OwlCap-7B. Furthermore, the InternVL2.5-8B variant attains the best Precision (40.5\%). These consistent and significant gains (+5.2\% to +7.3\% in F1) confirm that our self-evolving framework successfully reduces hallucinations and improves the overall faithfulness of detailed video captions.

\noindent\textbf{Alignment with human judgments.}
To strictly validate whether the scores assigned by our method accurately reflect caption quality, we conducted a pairwise preference evaluation using 500 randomly sampled caption pairs from the VDC-Agent-19K dataset. Each pair consists of a high-scoring caption ($y^{+}$) and a low-scoring caption ($y^{-}$). We invited three independent human evaluators to perform a blind review, alongside GPT-5.1 as an AI judge.

As presented in \cref{tab:human_gpt_eval}, the results demonstrate a \textit{strong consistency supporting the effectiveness of our method}. All three human evaluators consistently preferred $y^{+}$ over $y^{-}$, with preference rates ranging from 67.6\% to 75.8\%, significantly surpassing the cases where $y^{-}$ was favored. Furthermore, GPT-5.1 exhibits a stronger preference for our high-scoring captions, with an average of 89.8\%. Notably, we tested both input orders for the AI evaluator, and the consistent results confirm that the evaluation \textit{is robust to position bias}.

In addition to preference alignment, we further investigate \textit{potential length bias}. \cref{fig:length_distribution} illustrates the length distributions of the chosen captions ($y^{+}$) and the rejected captions ($y^{-}$). Although $y^{+}$ tends to be moderately longer on average, the two distributions overlap considerably. Moreover, the human evaluation in \cref{tab:human_gpt_eval} consistently favors $y^{+}$ in terms of semantic richness and visual grounding, further indicating that the observed preference is largely attributable to content quality rather than length alone.

\begin{table*}[t]
\scriptsize
\centering
\setlength{\tabcolsep}{1pt}
\caption{\textbf{Ablation on self-reflection and principle components.} Baseline is Qwen2.5-VL-7B-Instruct, \textbf{P} denotes Principle, and \textbf{R} denotes self-Reflection. We compare the baseline, naive principle-augmented input, test-time self-reflection, and our VDC-Agent. While adding principles helps, it increases inference cost and fails to generalize to unseen videos. Self-reflection further improves caption quality but is impractical at inference time. Our model achieves the best balance of accuracy and efficiency by internalizing self-reflection through training-time agentic data generation.}
\begin{tabular}{l|cc|cc|cc|cc|cc||cc|c}
\toprule
\multicolumn{1}{l|}{\multirow{2}{*}{Method}} & \multicolumn{2}{c|}{Camera} & \multicolumn{2}{c|}{Short} & \multicolumn{2}{c|}{Background} & \multicolumn{2}{c|}{Main Object} & \multicolumn{2}{c||}{Detailed} & \multicolumn{2}{c|}{Average} & \multicolumn{1}{c}{\multirow{2}{*}{Time}}\\
\multicolumn{1}{c|}{} & Acc. & Score & Acc. & Score & Acc. & Score & Acc. & Score & Acc. & Score & Acc. & Score & \\ \midrule
Baseline         & 42.61 & 2.18 & 37.76 & 1.91 & 44.10 & 2.22 & 49.00 & 2.47 & 46.25 & 2.35 & 43.95 & 2.23 & 15.5s\\
Baseline+P     & 47.76 & 2.43 & 38.81 & 1.96 & 47.82 & 2.41 & 50.07 & 2.53 & 47.87 & 2.42 & 46.47 & 2.35 & 22.3s \\
Baseline+P+R & 45.15 & 2.29 & 39.13 & 1.98 & 50.02 & 2.53 & 52.13 & 2.63 & 49.09 & 2.49 & 47.10 & 2.38 & 164.9s \\ \midrule
VDC-Agent        & \textbf{50.52} & \textbf{2.67} & \textbf{39.49} & \textbf{1.99} & \textbf{51.93}  & \textbf{2.62} & \textbf{53.23}  & \textbf{2.65}  & \textbf{50.21} & \textbf{2.55} & \textbf{49.08} & \textbf{2.50} & \textbf{15.5s} \\ \bottomrule
\end{tabular}
\label{tab:ab_agent}
\end{table*}

\begin{table*}[t]
\fontsize{6pt}{7pt}\selectfont
\centering
\setlength{\tabcolsep}{0.8pt}
\caption{\textbf{Ablation on curriculum strategies.} We compare fine-tuning strategies on the VDC benchmark. \textbf{Random DPO} shuffles data. \textbf{Anti-Curr. DPO} sorts by ascending $\Delta s$ (hard-to-easy). \textbf{Curr. DPO} sorts by descending $\Delta s$ (easy-to-hard, ours).}
\begin{tabular}{l|cc|cc|cc|cc|cc||cc}
\toprule
\multicolumn{1}{l|}{\multirow{2}{*}{Method}} &
  \multicolumn{2}{c|}{Camera} &
  \multicolumn{2}{c|}{Short} &
  \multicolumn{2}{c|}{Background} &
  \multicolumn{2}{c|}{Main Object} &
  \multicolumn{2}{c||}{Detailed} &
  \multicolumn{2}{c}{Average} \\
\multicolumn{1}{c|}{}      & Acc.  & Score & Acc.  & Score & Acc.  & Score & Acc.  & Score & Acc.   & Score & Acc.  & Score \\ \midrule
VDC-Agent (SFT)            & 45.43 & 2.36  & 39.30 & \textbf{1.99}  & 51.35 & 2.56  & 51.60 & 2.58  & 50.01 & 2.54  & 47.54 & 2.41  \\
VDC-Agent (Random DPO)     & 47.95 & 2.44  & 38.79 & 1.97  & 51.66 & 2.58  & 52.38 & 2.62  & 49.36 & 2.49  & 48.03 & 2.42  \\
VDC-Agent (Anti-Curr. DPO) & 47.31 & 2.42  & 39.20 & 1.98  & 51.69 & 2.58  & 51.19 & 2.56  & 48.86 & 2.47  & 47.65 & 2.40 \\
\rowcolor[HTML]{EFEFEF} 
VDC-Agent (Curr. DPO)      & \textbf{50.52} & \textbf{2.67}  & \textbf{39.49} & \textbf{1.99}  & \textbf{51.93} & \textbf{2.62}  & \textbf{53.23} & \textbf{2.65}  & \textbf{50.21} & \textbf{2.55}  & \textbf{49.08} & \textbf{2.50}  \\ \bottomrule
\end{tabular}
\label{tab:ab_dpo}
\end{table*}

\begin{table}[t]
\scriptsize
\centering
\caption{\textbf{Ablation on the robustness to principles.} Three contributors independently write principle sets (P1–P3), and P4 was derived from Shutterstock's industrial guidelines. R denotes self-reflection (please see \cref{tab:ab_agent}). The consistent performance across diverse principle sources demonstrates the robustness of VDC-Agent.}
\begin{tabular}{c|l|cc|ccc}
\toprule
\multirow{2}{*}{Contributor} & \multirow{2}{*}{Method} & \multicolumn{2}{c|}{VDCScore} & \multicolumn{3}{c}{DREAM-1K} \\
                             &                         & Acc.  & Score & F1   & Precision & Recall \\
\midrule
-                                & Baseline            & 43.95 & 2.23  & 30.1 & 30.5      & 29.7   \\
\midrule
\multirow{3}{*}{1}               & Baseline + P1       & 46.47 & 2.35  & 34.2 & 34.6      & 33.8   \\
                                 & Baseline + P1 + R   & 47.10 & 2.38  & 35.1 & 35.4      & 34.9   \\
                                 & VDC-Agent (with P1) & \textbf{49.08} & \textbf{2.50}  & \textbf{37.4} & \textbf{38.5}      & \underline{36.3}   \\
\midrule
\multirow{3}{*}{2}               & Baseline + P2       & 45.43 & 2.30  & 33.5 & 33.2      & 33.9   \\
                                 & Baseline + P2 + R   & 46.15 & 2.34  & 35.0 & 34.8      & 35.2   \\
                                 & VDC-Agent (with P2) & 48.84 & 2.48  & 36.9 & 37.1      & \textbf{36.8}   \\
\midrule
\multirow{3}{*}{3}               & Baseline + P3       & 46.85 & 2.37  & 33.7 & 34.1      & 33.4   \\
                                 & Baseline + P3 + R   & 47.25 & 2.39  & 34.5 & 34.9      & 34.1   \\
                                 & VDC-Agent (with P3) & \underline{49.02} & \textbf{2.50}  & \underline{37.1} & \underline{38.2}      & 36.0   \\
\midrule
\multirow{3}{*}{\shortstack{4\\(Industry)}} & Baseline + P4  & 46.10 & 2.33  & 33.8 & 34.0      & 33.6   \\
                                 & Baseline + P4 + R         & 46.90 & 2.37  & 34.8 & 35.1      & 34.5   \\
                                 & VDC-Agent (with P4)       & 48.95 & \underline{2.49}  & 36.6 & 37.3   & 35.9  \\
\bottomrule
\end{tabular}
\label{tab:ab_principle}
\end{table}

\begin{table}[t]
\scriptsize
\centering
\caption{\textbf{Ablation study of the maximum iteration $T$.} The generation time refers to the runtime for generating a sample on one GPU or the whole dataset on 8$\times$A800 GPUs. We set $T=4$ by default.}
\begin{tabular}{c|cc|ccc|c}
\toprule
\multirow{2}{*}{Max Iterations $T$} & \multicolumn{2}{c|}{VDCScore} & \multicolumn{3}{c|}{DREAM-1K} & \multirow{2}{*}{Dataset Construction Time} \\
                   & Acc.  & Score & F1   & Precision & Recall &                        \\ \midrule
2                  & 46.98 & 2.38  & 34.0 & 34.8      & 33.2   & 41.6s / 29.0h          \\
3                  & 48.15 & 2.43  & 36.7 & \underline{38.0}      & 35.5   & 58.3s / 40.6h          \\
4                  & 49.08 & 2.50  & \textbf{37.4} & \textbf{38.5}      & 36.3   & 70.7s / 49.2h          \\
5                  & \underline{49.29} & \underline{2.52}  & \underline{37.3} & 37.9      & \underline{36.8}   & 85.5s / 59.5h          \\
6                  & \textbf{49.55} & \textbf{2.53}  & 36.8 & 36.5      & \textbf{37.0}   & 103.8s / 72.2h         \\ \bottomrule
\end{tabular}
\label{tab:ab_t}
\end{table}

\subsection{Ablation and Discussion}
\noindent\textbf{Why is self-reflection effective?}
To better understand why our agentic self-reflection design is necessary and how each component contributes, we conduct ablation experiments on the VDC benchmark. We compare four variants: 
\begin{itemize}[nosep]
    \item \textbf{Baseline}: the original Qwen2.5-VL-7B-Instruct model without any additional mechanism; 
    \item \textbf{Baseline + Principle}: a naive variant that directly concatenates the textual principles into the prompt without agentic iteration; 
    \item \textbf{Baseline + Principle + Reflection}: which performs self-reflective refinement directly on the test set, but this leads to longer inference time.
    \item \textbf{VDC-Agent}: the proposed method.
\end{itemize}

As shown in \cref{tab:ab_agent}, simply injecting the principle text (Baseline+P) improves over the baseline, demonstrating that explicit quality criteria provide useful prior knowledge. However, this naive strategy not only increases inference time but also fails to adapt to diverse video inputs due to its fixed principle, resulting in only a modest +2.52 average accuracy gain. The third variant (Baseline+P+R) further enhances performance, validating that iterative diagnosis and correction indeed refine caption quality. Nevertheless, it assumes test-time adaptation~\cite{ding2025space} and significantly prolongs inference time, which is impractical for deployment. In contrast, our VDC-Agent achieves comparable or better performance without repeatedly accessing the test data. This demonstrates that VDC-Agent successfully distills the benefits of reflective refinement into a single forward model, providing both robustness and efficiency at inference time.

\noindent\textbf{Ablation on curriculum strategies.} 
To validate the effectiveness of our easy-to-hard scheduling, we compare four strategies in \cref{tab:ab_dpo}. Notably, Anti-Curriculum DPO results in performance degradation compared to Random DPO, suggesting that exposing the model to the most difficult samples prematurely hinders the alignment process. In contrast, our Curriculum DPO achieves the best performance across all metrics, confirming that the proposed easy-to-hard curriculum is essential for effective optimization.

\noindent\textbf{Robustness to principles.}
In our VDC-Agent, principles act as textual guidelines that steer automatic prompt optimization. A natural question is whether our performance depends on the exact wording or content of these principles. To verify the robustness, we conducted an ablation study to examine the \textit{sensitivity} to different principle sets. We invited three independent contributors to write their own versions of the principles, denoted as P1, P2, and P3, without mutual discussion. Additionally, to evaluate the applicability of our method to real-world industrial standards, we curated P4 derived from the official content review guidelines of Shutterstock~\cite{shutterstock_review}. These principles are detailed in the Appendix. Each set covers five task dimensions (camera, short, background, main object, and detailed) but differs in linguistic style, granularity, and emphasis. For each set, we re-ran the complete pipeline described in \S\ref{sec:method}. As shown in \cref{tab:ab_principle}, despite minor fluctuations across different sources, the final results remain highly consistent (around 49.0 accuracy and 2.5 score). This indicates that our improvement mainly stems from the agent’s iterative self-reflection and prompt refinement mechanism rather than the particular phrasing of principles, demonstrating strong robustness to principle design.

\noindent\textbf{Effect of the maximum iteration $T$.}
VDC-Agent’s refinement depth is controlled by the hyperparameter $T$, which caps the number of agent loops. As shown in \cref{tab:ab_t}, performance improves steadily from the baseline (no agent) to $T{=}4$, where accuracy and score largely saturate while keeping generation time reasonable. Increasing $T$ beyond $4$ yields marginal gains at the cost of additional latency, indicating a practical efficiency–performance trade-off. At the same time, this highlights our method’s compute flexibility: users with ample resources can obtain higher caption quality and more powerful captioner by increasing $T$, while $T{=}4$ remains a well-balanced default.
Moreover, the further comparison of overhead will be provided in the Appendix.

\begin{wraptable}{r}{0.5\textwidth}
\scriptsize
\centering
\caption{\textbf{Compute-normalized comparison.} Baseline: Qwen2.5-VL-7B. BoN: Best-of-N. P+R: Test-time Reflection.}
\label{tab:bon_comparison}
\begin{tabular}{lccc}
\toprule
Method               & Acc.  & Score & Time   \\
\midrule
Baseline (Greedy)    & 43.95 & 2.23  & 15.5s  \\
\rowcolor[HTML]{EFEFEF} 
Baseline (BoN, $N=6$)  & 44.88 & 2.28  & 93.0s  \\
\rowcolor[HTML]{DFDFDF} 
Baseline (BoN, $N=9$)  & 45.04 & 2.29  & 139.5s \\
\rowcolor[HTML]{CFCFCF} 
Baseline (BoN, $N=12$) & 45.12 & 2.30  & 186.0s \\
\rowcolor[HTML]{EFEFEF} 
Baseline+P+R ($T=2$)   & 46.53 & 2.35  & 87.4s  \\
\rowcolor[HTML]{DFDFDF} 
Baseline+P+R ($T=3$)   & \underline{46.86} & \underline{2.36}  & 131.7s \\
\rowcolor[HTML]{CFCFCF} 
Baseline+P+R ($T=4$)   & \textbf{47.10} & \textbf{2.38}  & 164.9s \\
\bottomrule
\end{tabular}
\end{wraptable}

\noindent\textbf{Compute-normalized comparison with Best-of-N.} 
To validate that our self-reflection mechanism's improvements derive from directed reasoning rather than simply scaling test-time compute~\cite{han2026unleashing,zhao2026shared}, we compare it against a Best-of-N (BoN) baseline under aligned computational budgets (average inference time per video). As shown in~\cref{tab:bon_comparison}, scaling $N$ in BoN yields marginal gains over the greedy baseline. Conversely, our Baseline+P+R ($T=4$) achieves 47.10\% accuracy and a 2.38 score in 164.9s, significantly outperforming BoN ($N=12$), which reaches only 45.12\% and 2.30 despite requiring more time (186.0s). This confirms that our agentic self-reflection provides a highly efficient, directed optimization path that is fundamentally superior to the undirected, trial-and-error approach of BoN.

\begin{figure*}[t]
  \centering
  \includegraphics[width=0.95\textwidth]{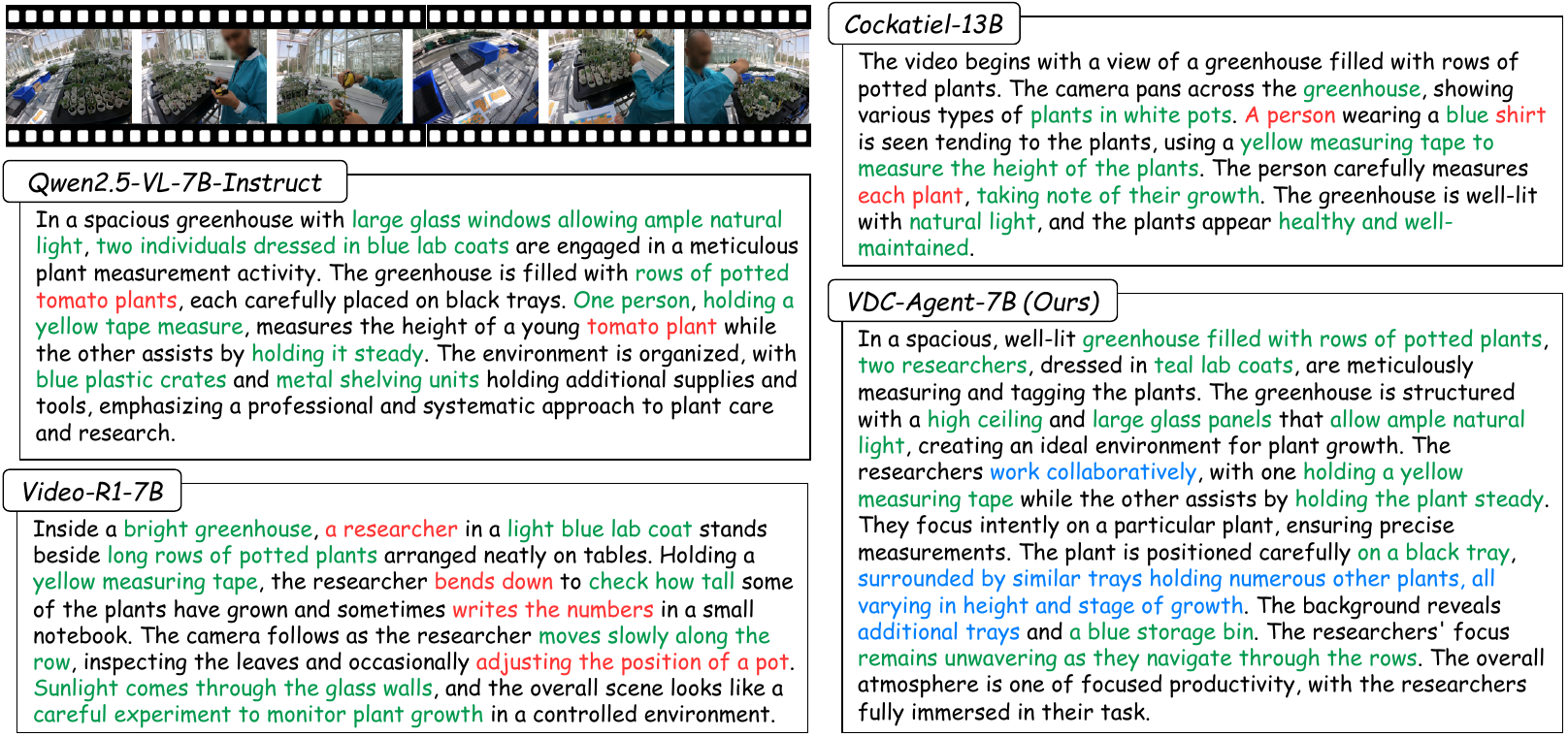}
  \caption{\textbf{Qualitative comparison on a video from the VDC benchmark.} We generate captions with different captioners under the same prompt: ``Please provide a detailed description of the given video.'' Text in green denotes \textcolor[HTML]{00994D}{correct, video-grounded} descriptions, while red indicates \textcolor[HTML]{FF3333}{incorrect or hallucinated} content. Blue highlights \textcolor[HTML]{007FFF}{fine-grained details} that are correctly identified only by our VDC-Agent-7B.}
  \label{fig:case}
\end{figure*}

\subsection{Qualitative Results}

\cref{fig:case} presents a qualitative comparison between VDC-Agent-7B and strong baselines on a greenhouse video. While Qwen2.5-VL-7B recognizes the plant-measuring activity, it hallucinates unsupported details (\emph{e.g.}, ``tomato plants''). Video-R1-7B and Cockatiel-13B provide more elaborate descriptions but omit crucial background elements (\emph{e.g.}, tray arrangements) and camera motion (\emph{e.g.}, the smooth pan along the aisle as the researchers walk and measure the plants). 

In contrast, VDC-Agent-7B produces a more structured and comprehensive narrative. It better captures scene layout, entity interactions, and the activity's intent, alongside depicting subtle attributes like spatial organization and camera motion. These observations align with our quantitative results, confirming that agentic self-reflection enables the generation of richer, more faithful, and better-grounded detailed captions.


\section{Conclusion}
We propose VDC-Agent, which upgrades off-the-shelf MLLMs for detailed video captioning via iterative self-reflection. By leveraging self-generated preference pairs ordered by difficulty, our approach effectively aligns models to produce precise descriptions. VDC-Agent-7B establishes a new state-of-the-art, outperforming prior baselines while maintaining efficient inference.

\noindent\textbf{Future Work.} We will extend our approach to larger backbones (\emph{e.g.}, Qwen-32B) and broader video understanding tasks (\emph{e.g.}, video question answering) to further investigate the performance ceilings, scalability, and universality of the proposed agentic framework.

\section*{Acknowledgements}
\begin{sloppypar}
This work is supported by the National Natural Science Foundation of China under Grant 62576224 and 62302382, the Guangdong Provincial Key Laboratory of Computility Microelectronics 2024B1212010007, the Guangdong Basic and Applied Basic Research Foundation Project 2026A1515011733, the Shenzhen Key Technical Projects under Grant CJGJZD20220517141605013, JCYJ20220818101406014 and JSGG20220831105801004, China Postdoctoral Science Foundation 2024M752584, and Natural Science Foundation of Shaanxi Province 2024JC-YBQN-0637.
\end{sloppypar}

%
%
\bibliographystyle{splncs04}
\bibliography{main}

\end{document}